\pdfoutput=1

\documentclass[11pt]{article}
\usepackage{booktabs}
\usepackage{graphics,graphicx,caption,float,color}
\usepackage[utf8]{inputenc} 
\usepackage[T1]{fontenc}    
\usepackage[hidelinks]{hyperref}       
\usepackage{url}            
\usepackage{booktabs}       
\usepackage{amsfonts,amssymb}       
\usepackage{nicefrac}       
\usepackage{microtype}      
\usepackage{xcolor}         
\usepackage{colortbl}
\usepackage{bigstrut}
\usepackage{pifont}
\usepackage{graphicx}
\usepackage{amsmath}
\usepackage{amssymb}
\usepackage{amsthm}
\usepackage{caption}
\usepackage{subfig}
\usepackage{svg}
\usepackage{multirow}
\usepackage{float}
\usepackage{wrapfig}
\usepackage{acl}
\usepackage{times}
\usepackage{latexsym}

\usepackage[T1]{fontenc}

\usepackage[utf8]{inputenc}

\usepackage{microtype}

\usepackage{inconsolata}
\usepackage{xspace}
\usepackage{ulem}
\newcommand{\llamapro}{\textsc{LLaMA Pro}\xspace}
\newcommand{\instruct}{\textsc{LLaMA Pro - Instruct}\xspace}

\definecolor{LightCyan}{rgb}{0.88,1,1}
\definecolor{lightgoldenrodyellow}{rgb}{0.98, 0.98, 0.82}

\title{\includegraphics[scale=0.05, bb=-100 8 300 34]{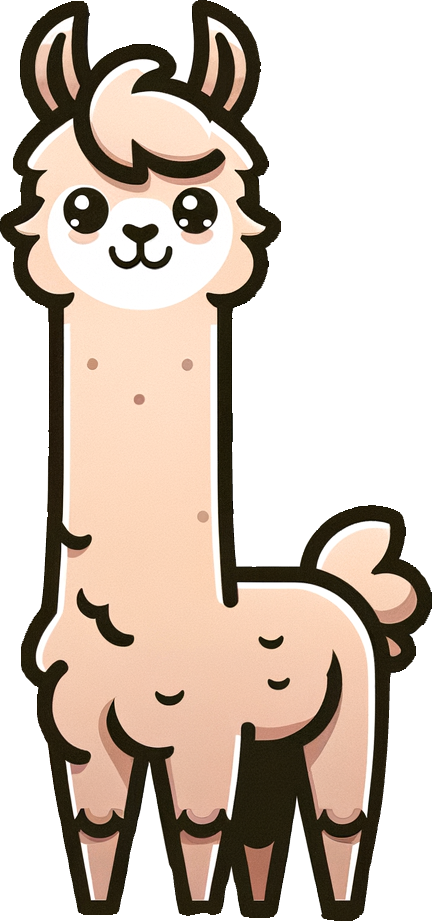}~~ \llamapro: Progressive LLaMA with Block Expansion}

\author{Chengyue Wu$^{1,2}$ \qquad Yukang Gan$^{2}$ \qquad Yixiao Ge$^{2}$\thanks{Correspondence to \href{mailto:yixiaoge@tencent.com}{\texttt{yixiaoge@tencent.com}}.} \\
    {\bf Zeyu Lu}$^{3}$ \qquad {\bf Jiahao Wang}$^{1}$ \qquad {\bf Ye Feng}$^{4}$ \qquad {\bf Ying Shan}$^{2}$ \qquad {\bf Ping Luo}$^{1}$ 
 \\~\\
  $^1$The University of Hong Kong \qquad $^2$ARC Lab, Tencent PCG \\
  $^3$Shanghai Jiao Tong University \qquad $^4$Beijing Language and Culture University \\~\\
  \url{https://github.com/TencentARC/LLaMA-Pro}
  }

\begin{document}
\maketitle

\begin{abstract}

Humans generally acquire new skills without compromising the old; however, the opposite holds for Large Language Models (LLMs), \textit{e.g.}, from LLaMA to CodeLLaMA.
To this end, we propose a new post-pretraining method for LLMs with an expansion of Transformer blocks.
We tune the expanded blocks using only new corpus, efficiently and effectively improving the model's knowledge while mitigating forgetting.
In this paper, we experiment on the corpus of code and math, yielding \textbf{\llamapro-8.3B}, a versatile foundation model initialized from LLaMA2-7B, excelling in general tasks, programming, and mathematics. 
%
%
%
%
\llamapro and its instruction-following counterpart (\instruct) achieve advanced performance among various benchmarks,
demonstrating superiority over existing open models in the LLaMA family and the immense potential of reasoning and addressing diverse tasks as an intelligent agent.
Our findings provide valuable insights into integrating natural and programming languages, laying a solid foundation for developing advanced language agents that operate effectively in various environments.
\end{abstract}

\begin{figure}[t]
    \centering
    \includegraphics[width=\linewidth]{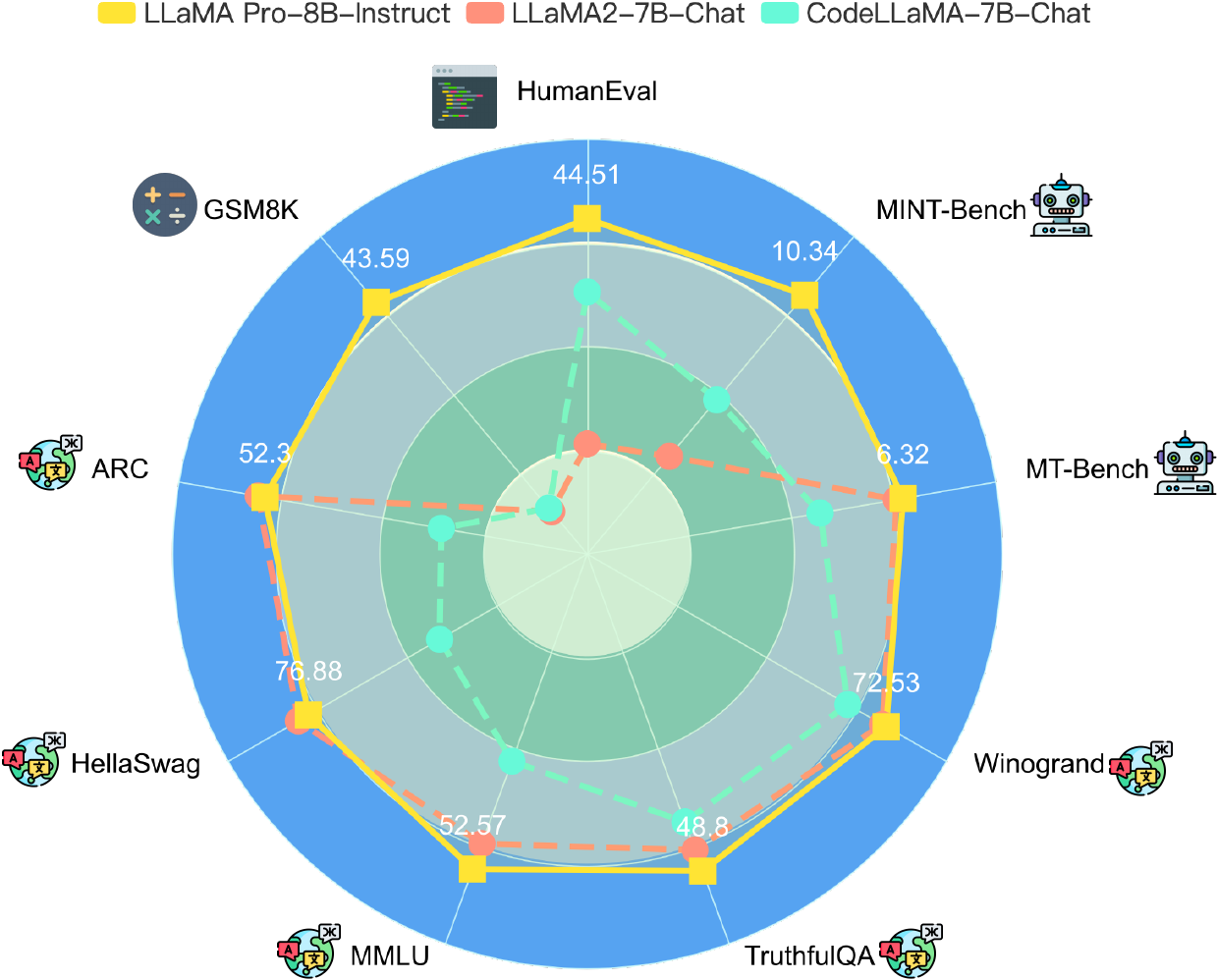}
    \caption{
         \instruct delivers state-of-the-art performance across a wide variety of tasks, ranging from general language to specific domains, superior to existing models from the LLaMA series.
    }
    \label{fig:radar}
\end{figure}

\section{Introduction}

The advent of Large Language Models (LLMs) has revolutionized the field of natural language processing, exhibiting remarkable proficiency in a variety of real-world tasks~\cite{gpt4,chowdhery2023palm}. 
Despite the versatility, LLMs still fall short in certain domains, for example, programming, mathematics, biomedical, or finance.
This limitation impedes the progress of developing generic language agents for broader applications.

Existing works \cite{liu2023llm360,li2023starcoder,wu2023bloomberggpt} attempted to improve the multi-faceted capabilities of pre-trained LLMs with tailored data recipes. 
While feasible, they require substantial computational resources and vast amounts of data, which poses a challenge to the democratization of LLM research.
Consequently, another line of research, known as domain-adaptive pretraining, focuses on post-pretraining with domain-specific corpora~\cite{gururangan2020don}. These approaches have demonstrated efficacy in adapting various LLMs to specific domains~\cite{roziere2023code,azerbayev2023llemma,wu2023bloomberggpt,xu2023lemur}, 
resulting in enhanced performance on downstream domain-specific tasks at a reduced computational cost.



Nonetheless, a considerable obstacle emerges in catastrophic forgetting~\cite{de2021continual}. Post-pretraining often leads to a decline in the model's original general abilities, inhibiting the fine-tuned performance of the model on diverse tasks~\cite{cheng2023adapting,dong2023abilities}. This necessitates a method that can inject domain-specific knowledge into LLMs while preserving their general abilities, thereby enhancing their comprehensive capabilities.

\begin{figure*}[t]
    \centering
    \includegraphics[width=0.8\linewidth]{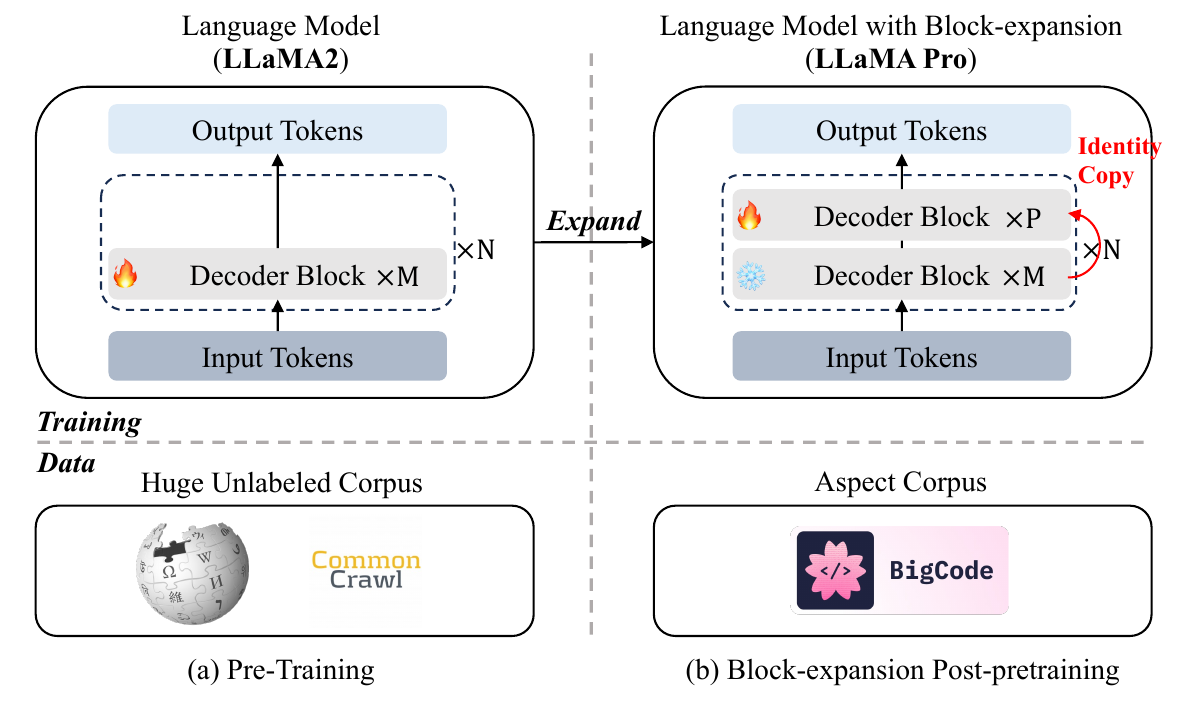}
    \caption{
         (a) We begin with a large language model (LLM) pre-trained on a massive unlabeled corpus, resulting in a model with strong general capabilities. Here we select the off-the-shelf LLaMA2 for convenience.
         (b) We employ backbone expansion and fine-tune the expanded identity blocks using the aspect corpus while freezing the blocks inherited from the base model. The model after post-pretraining can be used for instruction tuning as usual.
    }
    \vspace{-0.3cm}
    \label{fig:pipeline}
\end{figure*}

Towards this end, we introduce a simple yet effective post-pretraining method, termed \textit{block expansion}. 
We expand the off-the-shelf pre-trained LLM using copied Transformer blocks, as illustrated in Figure \ref{fig:pipeline}. 
The newly added blocks, whose linear layers are zero-initialized to enable identity mapping, are further tuned with only domain-specific corpus while the remaining blocks are frozen.
After tuning, the extended pre-trained model excels in both general and domain-specific tasks.


In practice, we extend the pre-trained LLaMA2-7B~\cite{touvron2023llama} by eight more blocks, yielding \llamapro, a foundation model with 8.3B parameters, and enhanced performance in programming, coding, and reasoning.
We pre-train \llamapro's expanded blocks on 80B tokens using open-source code and math data for 2830 GPU Hours (16 NVIDIA H800 GPUs for about 7 days).
We further perform supervised instruction tuning (fully fine-tuning of all the blocks, \textit{aka} SFT) on \llamapro with approximately 80M tokens, yielding \instruct.
It is noted that pre-trained models produced by our block expansion method are well-compatible with the subsequent SFT techniques without specific modification.


As shown in Figure~\ref{fig:radar}, \instruct reaches state-of-the-art performance across a broad range of general, code (\textit{i.e.}, HumanEval), and math (\textit{i.e.}, GSM8K) tasks. 
Furthermore, we assess the capabilities of \instruct as a language agent across various scenarios (\textit{i.e.}, MINT-Bench), with a focus on the tool usage abilities and the capacity to ground in environmental and human feedback. 
We also employ GPT-4~\cite{gpt4} automatic evaluation to assess \llamapro's ability to serve as an effective assistant (\textit{i.e.}, MT-Bench). 
Comprehensive experimental results indicate the superiority of \instruct over other models from the LLaMA family on both benchmarks and practical applications.
%
Our contributions are three-fold:
\begin{itemize}
    \item We propose a novel post-pretraining method for LLMs, termed block expansion, enabling the injection of new knowledge while preserving the initial capabilities.
    \item We introduce \llamapro and \instruct, versatile LLMs that well integrate natural and programming languages, excelling in general tasks, programming, and mathematics.
    \item We benchmark the family of \llamapro on extensive datasets, including both traditional and agent-oriented tasks, demonstrating its superiority and great potential in broader complex applications.
\end{itemize}
\section{Related Work}
\paragraph{Advancements in Large Language Models.}
Recent advancements in large language models have led to significant progress, with model and data scale growth driving state-of-the-art performance across various tasks~\cite{hoffmann2022training, kaplan2020scaling, chowdhery2023palm}. The development of generalist models has enabled addressing diverse problems and rapid adaptation to new tasks~\cite{radford2019language, brown2020language}. The open-source community has further contributed by releasing powerful models like LLaMA~\cite{touvron2023llama} and CodeLLaMA~\cite{roziere2023code}. Our work builds upon these developments, providing a method for specializing LLMs in the code domain, fostering future research and applications.
\paragraph{Post-pretraining.}
Language model applications typically involve a two-step process: general-domain pretraining followed by domain-specific training~\cite{roziere2023code,azerbayev2023llemma}. Fine-tuning often aims to enhance instruction-following abilities~\cite{sanh2021multitask, wei2021finetuned,wang2023far} or align model outputs with human preferences~\cite{ziegler2019fine,ouyang2022training, bai2022training}. Some research explores parameter-efficient fine-tuning methods for adapting pretrained models to new domains~\cite{houlsby2019parameter, hu2021lora, pmlr-v202-wu23t}, while others focus on continual learning post-pretraining~\cite{wang2023trace,gupta2023continual,scialom2022fine}.
Parameter-efficient tuning methods like adaptor and LoRA are generally applied during the instruction tuning phase rather than the pretraining phase. In contrast, our focus is on enhancing the capacity of LLMs by increasing their depth during continued pretraining. Our work proposes an adaptation strategy that combines continued training with general capability maintenance, allowing LLMs to specialize without sacrificing overall performance.
\paragraph{Progressive Learning.}
Progressive training has gained attention for accelerating large-scale model training in computer vision~\cite{zhang2023adding} and NLP research~\cite{yao20232x,li2023flm}. \citet{gong2019efficient} proposed a stacking method doubling model depth successively. CompoundGrow\cite{gu2020transformer} extends stacking with Feed-Forward Network expansion in schedule design. \citet{shen2022staged} introduced a staged method supporting hidden size expansion. Bert2BERT\cite{chen2021bert2bert} and LiGO~\cite{wang2023learning} accommodate all growth dimensions. Our method utilizes depth growth to maintain general performance while adapting to specific domains.
\section{Method}

\subsection{Preliminaries: The LLaMA Block}
\begin{figure}[t]
    \centering
    \includegraphics[width=\linewidth]{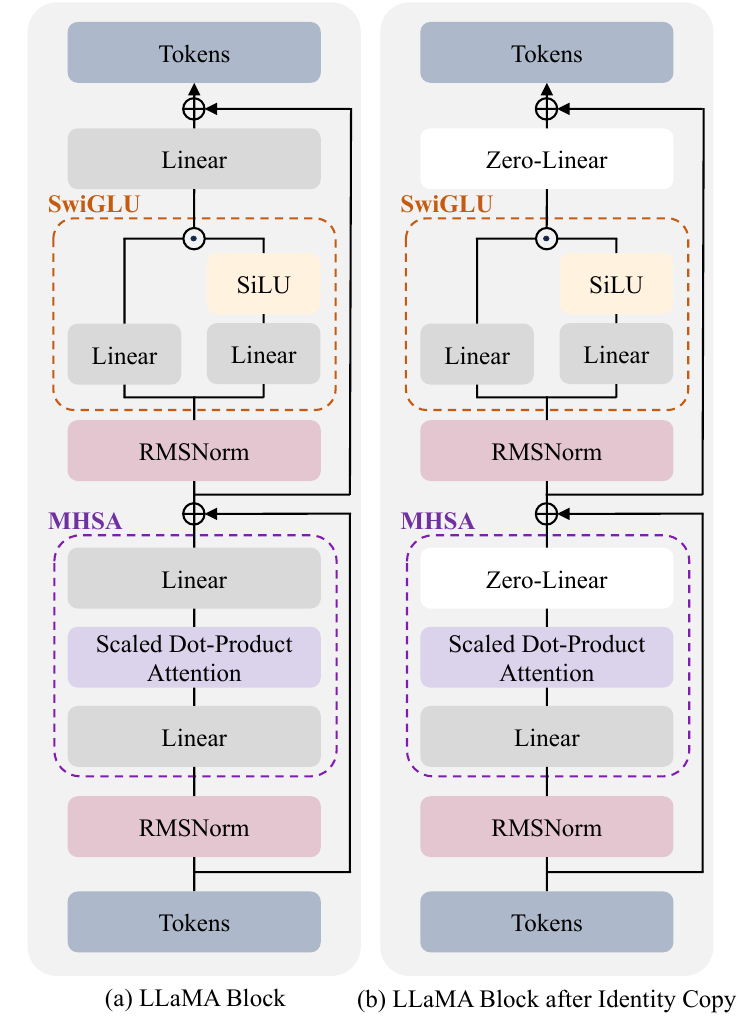}
    \caption{
         (a) An overview of the LLaMA Block, comprising an MHSA mechanism followed by the FFN with SwiGLU activation. (b) The Identity LLaMA block after an identity copy, achieved by initializing the output linear matrix to zero in order to preserve the output from the base LLaMA model.
    }
    \vspace{-0.3cm}
    \label{fig:identical_copy}
\end{figure}

The LLaMA block consists of a multi-head self-attention (MHSA) mechanism followed by a position-wise feed-forward network (FFN) with residual connections and a Swish-Gated Linear Unit (SwiGLU) operation as Figure~\ref{fig:identical_copy} shows. Given an input $x$, the LLaMA block produces an output $y$ as described by the following equations:
\begin{equation}\label{eq:llama_block}
\begin{split}
x' = x + \text{MHSA}(\text{RMSNorm}(x)) \\
y = x' + \text{FFN}(\text{RMSNorm}(x'))
\end{split}
\end{equation}
The input $x$ has a dimension of $n \times d$, where $n$ is the sequence length and $d$ is the hidden size. The output $y$ has the same dimension as the input $x$. The MHSA operation is a crucial component of the transformer, defined as:
\begin{equation}
\text{MHSA}(Q, K, V) = \text{Concat}(\text{head}_1, \dots, \text{head}_h)W^O
\end{equation}
where $Q$, $K$, and $V$ are the query, key, and value matrices, respectively, and $W^O$ is the output weight matrix without bias    . Each head is computed as:
\begin{equation}
\begin{split}
\text{head}_i = \text{Attention}(xW^Q_i, xW^K_i, xW^V_i) \\
\text{Attention}(Q_i, K_i, V_i) = \text{Softmax}\left(\frac{Q_iK_i^T}{\sqrt{d_k}}\right)V_i
\end{split}
\end{equation}
with $W^Q_i$, $W^K_i$, and $W^V_i$ being the corresponding weight matrices for the $i$-th head.

The FFN block in the LLaMA model utilizes the SwiGLU activation function, which is defined as:
\begin{equation}
\begin{split}
\text{SwiGLU}(x, W, V) = \text{SiLU}(x W) \otimes (x V) \\
\text{FFN}(x) = \text{SwiGLU}(x, W_1, W_2)W_3
\end{split}
\end{equation}
where $\otimes$ denotes element-wise multiplication, $W_1$, $W_2$, and $W_3$ are the weight matrices without bias, $\text{SiLU}(x) = x \otimes \sigma(x)$.

\subsection{Block Expansion}
Given a model with blocks $(\phi_0, \phi_1, . . ., \phi_L)$, the block expansion incorporates an identity block $\phi_{id}$ after each block in the original model, ensuring that the expanded model maintains the same output after expansion. The identity block is defined as $\phi_{id}(x) = x$ so the input and output are identical.

Suppose we have an initial model with $L$ blocks that needs to be expanded to $L'$ blocks. First, we partition the original $L$ blocks into $N$ groups, with each group containing $\frac{L}{N}$ blocks. For each group, we create identity copies of the top $P$ blocks and stack them on top of each group, as depicted in Figure \ref{fig:identical_copy}. We arrange these blocks in an interleaved manner to maintain the structural characteristic of the transformer model, whose prior is that deeper blocks encode more complex information ~\cite{van2019does,tenney2019bert}. This process leads to an increased depth in the model while maintaining its output behavior.

Shen et al.~\cite{shen2022staged} proposed the initialization of scale parameters in the Norm modules within the identity blocks to zero for the construction of the identity block. However, this approach may not be effective when applied to the LLaMA block. The reason lies in the fact that the gradient of the loss function $L$ with respect to the RMSNorm weight $w$ during backpropagation would be zero. This would prevent the training of RMSNorm, implying that when $\text{RMSNorm}(x') = 0$, the following condition will hold:
\begin{equation}
\frac{\partial L}{\partial w} = \frac{\partial L}{\partial y} \frac{\partial \text{FFN}(\text{RMSNorm}(x'))}{\partial \text{RMSNorm}(x')} \frac{\partial \text{RMSNorm}(x')}{\partial w} = 0.
\end{equation}
This equation signifies that the gradient of the loss function with respect to the weight of RMSNorm is zero, which would hinder the training of the RMSNorm module. This is further explained in Appendix~\ref{sec:appendix_gradient}. Referring to the LLaMA block formulation in Equation~\ref{eq:llama_block}, the identity can be achieved as long as $\text{MHSA}(\text{RMSNorm}(x))=\mathbf{0}$ and $\text{FFN}(\text{RMSNorm}(x'))=\mathbf{0}$. We initialize the $W^O$ and $W_3$ weight matrices in the identity blocks to zero. Due to the presence of residual connections and the absence of bias terms in the LLaMA block, only the residual flows through the identity block. As a result, the entire block is reduced to an identity block at initialization, preserving the output from the initial model.

The entire training pipeline is depicted in Figure~\ref{fig:pipeline}. Our method concentrates on the post-pretraining stage, targeting specific domain corpora. We begin by initializing our model with large language models trained on extensive unlabeled general corpora, where all blocks will be fine-tuned. To enhance the model's capacity for accommodating additional domain knowledge while retaining its general knowledge, we employ block expansion to increase the number of blocks in the LLM. During this process, we only fine-tune the newly added blocks while freezing the original blocks, thereby preserving the general abilities of the model.

\section{Experiments}
This section presents our key experimental findings. We begin with experimental settings (described in Sec.~\ref{sec:experimental_setting}), and then verify the effectiveness of block expanded tuning after pretraining (described in Sec.~\ref{sec:pretraining}). Next, we give the supervised finetuning (SFT) results (described in Sec.~\ref{sec:sft}).  Finally, ablation studies of the key design choices are presented (described in Sec.~\ref{sec:ablation}).
\subsection{Experimental Settings}
\paragraph{Pretrain details.} We construct a dataset that concentrates on code and math. For the code component, we rely on the Stack-dedup dataset, which is a compilation of permissively licensed source codes from GitHub. 
Among all the programming languages available in Stack-dedup, we specifically utilize the Python split. 
As for the math component, we opt for the Proof-pile-2 dataset~\cite{azerbayev2023llemma}, a 55-billion-token amalgamation of scientific papers, web data containing mathematical content, and mathematical code. The details can be found in Appendix \ref{sec:appendix_data}.

We initialize our base model with LLaMA2-7B and expand the number of blocks from 32 to 40 using an interleaved approach. In the block expansion process, we configure the parameters as $P=1$, $M=4$, and $N=8$, resulting in 8 groups where each group expands from 4 blocks to 5 blocks. For the code and math corpus pretraining, we employ a batch size of 1024, a sequence length of 4096, a warmup ratio of 6\%, a learning rate of 2e-4, and a Cosine learning rate scheduler. We also use bf16 mixed precision, a weight decay of 0.1, and gradient clipping at 1.0. To speed up the training process, we apply the flash-attention mechanism.

Our experiment is conducted on 16 NVIDIA H800 GPUs. \llamapro is trained for a total of 15,900 steps. This training process corresponds to approximately 2830 H800 GPU hours.

We want to highlight that our approach does not incur higher training costs, and it is worth the extra resources to achieve a better performance of the domain specific tasks in the inference.

\textbf{Training stage cost:} Our approach requires fewer computational resources since only the newly added blocks are tuned during training. As illustrated in Figure ~\ref{fig:scatter}, LLaMA Pro-8B (1B parameters tuned for 80B tokens) incurs less training overhead compared to CodeLLaMA-7B (7B parameters tuned for 500B tokens). It also uses fewer resources than training domain-specific models from scratch, such as StarCoder and CrystalCoder. Despite this, our method achieves a better balance of general and domain-specific performance, offering a more cost-effective solution.

\textbf{Inference stage cost:} Although our method requires more resources during inference than the initial LLM, it strikes a balance between performance and efficiency. LLaMA Pro-8B outperforms larger models like LLaMA2-13B and LLaMA2-34B in the code domain while demanding significantly fewer resources during training and inference.

\paragraph{SFT details.}
During the instruction fine-tuning phase, we combine five data sources to create \instruct as shown in Table \ref{tab:instruction-tuning-datasets}. The final sft dataset consists of approximately 1M samples. To fine-tune the basic models, we employ specific configurations, including a batch size of 128, a sequence length of 4096, 0.03 warmup ratio, a learning rate of 2e-5, a Cosine learning rate scheduler, and bf16 mixed precision.

\paragraph{Evaluation details.} We conduct a comparative analysis of \llamapro with the latest state-of-the-art (SOTA) Large Language Models (LLMs). The evaluation is performed on six key general benchmarks using the Eleuther AI Language Model Evaluation Harness\footnote{\url{https://github.com/EleutherAI/lm-evaluation-harness}}, a unified framework designed to test generative language models across a vast array of evaluation tasks. For code-related tasks, we employ the BigCode Evaluation Harness\footnote{\url{https://github.com/bigcode-project/bigcode-evaluation-harness}} to evaluate HumanEval and MBPP, and we report the pass@1 rate of code tasks with greedy decoding. The evaluation details can be found in Appendix \ref{sec:appendix_evaluation}.

\label{sec:experimental_setting}
\subsection{Pretrain Results}
\begin{table*}[t]
\resizebox{1.\linewidth}{!}
{
\begin{tabular}{lccccc|cc|cc|c}
\toprule
\textbf{Model} & \multicolumn{5}{c|}{\textbf{Language Tasks}} & \multicolumn{2}{c|}{\textbf{Math Tasks}} & \multicolumn{2}{c|}{\textbf{Code Tasks}} & \textbf{Avg.}  \bigstrut\\
& ARC & HellaSwag & MMLU & TruthfulQA & Winogrande & GSM8K & GSM8K-PoT & HumanEval & MBPP & \\
\midrule
\multicolumn{11}{l}{\textit{Pretrained comparison}} \\
\rowcolor{lightgoldenrodyellow}
\llamapro (8B) & 54.10 & 77.94 & 47.88 & 39.04 & 73.95 & 17.89 & 25.42 & 28.66 & 33.20 & \textbf{44.23}  \bigstrut\\
CrystalCoder (7B) & 47.01  & 71.97 & 48.78 & 35.91 & 67.17 & 10.77 & 24.96 & 28.38 & 36.38 & 41.26 \bigstrut\\
LLaMA2-7B & 53.07 & 78.59 & 46.87 & 38.76 & 74.03 & 14.48 & 17.68 & 13.05 & 20.09 & 39.62 \bigstrut\\
CodeLLaMA-7B & 39.93 & 60.80 & 31.12 & 37.82 & 64.01 & 5.16 & 25.20 & 33.50 & 41.40 & 37.66 \bigstrut\\
StarCoder-15B & 30.38 & 47.93 & 29.96 & 41.28 & 56.12 & 9.48 & 25.09 & 33.63 & 43.28 & 35.24 \bigstrut\\
LLaMA-7B & 50.94 & 77.81 & 35.69 & 34.33 & 71.43 & 8.04 & 10.46 & 10.61 & 17.04 & 35.15 \bigstrut\\
OpenLLaMA-v2-7B & 43.69  & 72.20 & 41.29 & 35.54 & 69.38 & 3.49 & 5.46 & 15.32 & 12.69 & 33.23 \bigstrut\\
Falcon-7B & 47.87  & 78.13 & 27.79 & 34.26 & 72.38 & 4.62 & 4.32 & 9.42 & 13.39 & 32.46 \bigstrut\\
\midrule
\multicolumn{11}{l}{\textit{SFT comparison}} \\
\rowcolor{lightgoldenrodyellow}
\instruct & 52.30 & 76.88 & 52.57 & 48.80 & 72.53 & 43.59 & 55.61 & 44.51 & 37.88 & \textbf{53.85} \bigstrut\\
LLaMA2-7B-Chat & 52.90 & 78.55 & 48.32 & 45.57 & 71.74 & 7.35 & 19.73 & 14.63 & 21.60 & 40.04 \bigstrut\\
CodeLLaMA-7B-Instruct & 36.52 & 55.44 & 34.54 & 41.25 & 64.56 & 7.96 & 34.67 & 34.80 & 44.4 & 39.35 \bigstrut\\
WizardCoder-Python-7B & 41.81 & 65.06 & 32.29 & 36.32 & 61.72 & 4.70 & 17.60 & 42.07 & 47.20 & 38.75\bigstrut\\
WizardMath-7B & 54.10 & 79.55 & 45.97 & 43.65 & 72.69 & 2.73 & 25.57 & 12.20 & 18.00 & 39.38\bigstrut\\
\bottomrule
\end{tabular}}
\caption{Comparison of evaluation results among several prominent code and language models.}
\label{tab:llm-comparison}
\end{table*}
\begin{figure}[t]
    \centering
    \includegraphics[width=\linewidth]{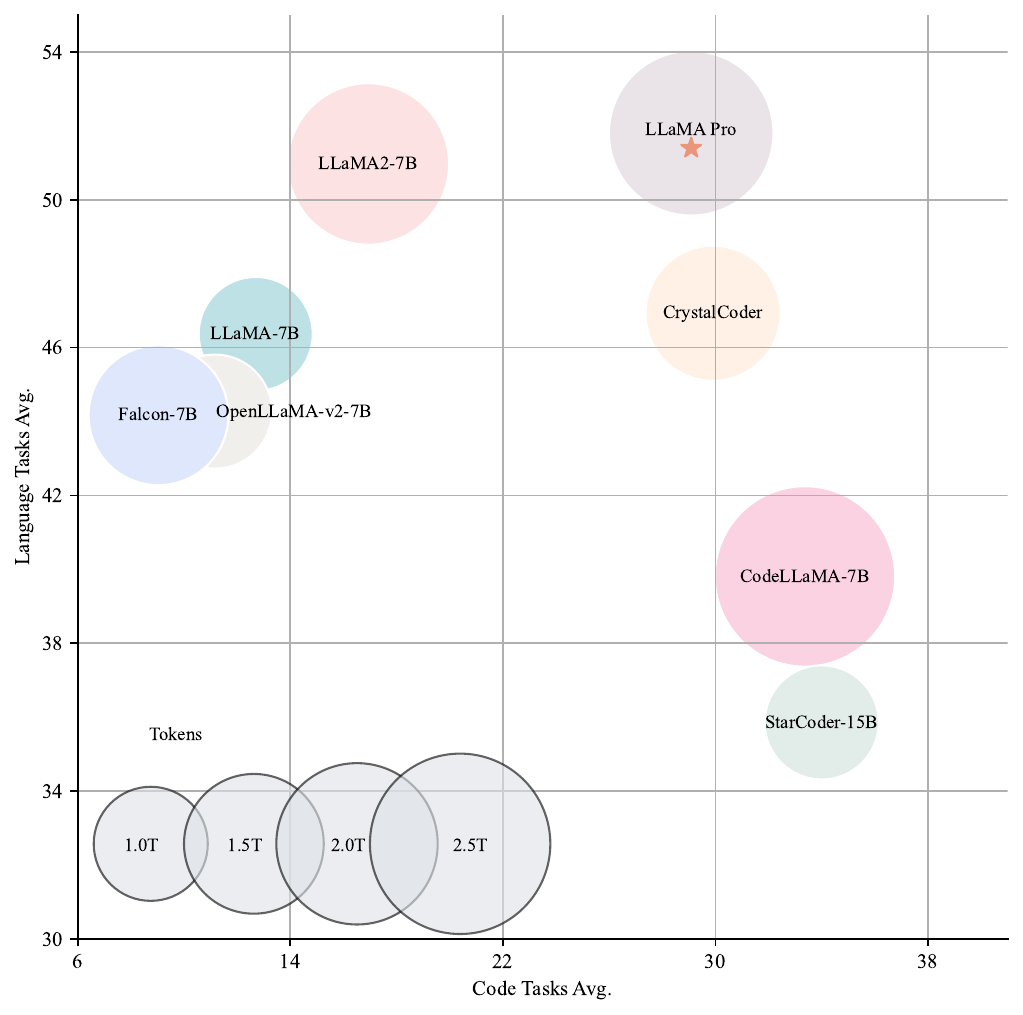}
    \caption{
         We compare \llamapro's general performance and code performance to a set of models trained around the same time, spanning from general LLMs to code-oriented LLMs. The size of the blobs is proportional to the number of tokens trained. Mistral-7B is not included here, as the number of tokens is not reported in its paper.}
    \label{fig:scatter}
\end{figure}

We evaluate \llamapro's performance with benchmark datasets from the Open LLM Leaderboard. Furthermore, we incorporate coding benchmark datasets, including HumanEval pass@1 and MBPP pass@1, as well as the math benchmark GSM8K, to provide a comprehensive evaluation. We compare the performance of \llamapro with a selection of state-of-the-art pretrained models that were trained around the same period with similar size. This includes general-purpose pretrained models like LLaMA2 and code-oriented pretrained models like CodeLLaMA. The results are presented in Table~\ref{tab:llm-comparison}.

The results highlight that \llamapro effectively balances natural language processing and coding capabilities. It not only preserves the general performance of its base model, LLaMA2-7B, but also surpasses it in the average performance of general language tasks. Conversely, CodeLLaMA-7B sacrifices general performance. We attribute this improvement to our expansion design, which freezes the initial LLaMA blocks to maintain their capabilities and increases the blocks to accommodate domain-specific knowledge.

As depicted in Figure~\ref{fig:scatter}, \llamapro shows robust general performance alongside code performance that is on par with code-oriented LLMs. Situated on the Pareto frontier, \llamapro has undergone fine-tuning with an additional 80B tokens in conjunction with LLaMA2, which more than doubles the code tasks average performance. In contrast, CodeLLaMA is fine-tuned with 500B tokens. \llamapro excels in general performance while maintaining code performance that is competitive with code-oriented LLMs, whether they are trained from scratch, such as StarCoder-15B and CrystalCoder, or fine-tuned like CodeLLaMA-7B.

\label{sec:pretraining}

\subsection{SFT Results}

\begin{table}[t]
\centering
\begin{tabular}{lc}
\toprule
\textbf{Model} & \textbf{MT Bench}\\
\midrule
Alpaca-13B & 4.53 \\
CodeLLaMA-7B-Instruct & 5.71 \\
Vicuna-7B & 6.17 \\
LLaMA2-7B-Chat & 6.27 \\
\rowcolor{lightgoldenrodyellow} \instruct & \textbf{6.32} \\
\bottomrule
\end{tabular}
\caption{GPT-4 automatic evaluation of Chatbot models. \instruct outperforms widely used LLaMA community chatbots.}
\label{tab:mt_bench}
\vspace{-3mm}
\end{table}
Modern LLMs typically undergo supervised fine-tuning or instruction tuning after pretraining on vast amounts of unlabeled data. In this section, we aim to demonstrate that our expansion strategy can adapt to this widely used training pipeline, just as traditional LLMs do.

Table~\ref{tab:llm-comparison} presents a comparison of evaluation results among several prominent supervised fine-tuning (SFT) LLMs from the LLaMA community, across general tasks, math tasks, and code tasks benchmarks. As a singular SFT model, \instruct attains state-of-the-art performance, even when compared to specifically tuned models such as WizardCoder and WizardMath. This demonstrates its more comprehensive capabilities.

As seen in Figure~\ref{fig:radar}, \instruct boosts both code and math tasks to SOTA performances while maintaining reliable general performance. We enhance the average performance of LLaMA2-7B-chat and CodeLLaMA-7B-instruct by 13.81\% and 14.50\% respectively, which highlights the benefits of balancing textual and coding abilities.

To assess the comprehensive conversational performance of the \instruct assistant, we evaluate it using the MT-Bench with GPT-4 automatic scoring, as proposed by Vicuna~\cite{zheng2023judging}. As depicted in Table~\ref{tab:mt_bench}, \instruct surpasses widely used chatbots from the LLaMA community. This indicates its potential as a chatbot capable of providing helpful responses, in addition to its impressive performance in traditional benchmarks. The details of MT-Bench can be found in the Appendix \ref{sec:appendix_mt}.

We use MINT-Bench ~\cite{wang2023mint} to evaluate our model's ability to solve multi-turn interactions by using tools. MINT-Bench tests LLMs' ability to use tools by generating and executing Python code, focusing on tool-augmented task-solving and leveraging natural language feedback. MINT includes eight datasets covering reasoning, code generation, and decision-making. The details of MINT can be found in the Appendix \ref{sec:appendix_mint}. The results are shown in Table~\ref{tab:mint}. \instruct achieves SOTA performance compared to similar size models in multi-turn interactions with the use of tools.
\begin{table}[t]
\centering
\resizebox{\linewidth}{!}
{\scriptsize
\begin{tabular}{c|ccccc|c}
\toprule
\textbf{Model} & \multicolumn{5}{c|}{\textbf{Interaction Turns}} & \textbf{Avg.}  \bigstrut\\
& 1 & 2 & 3 & 4 & 5  \\
\midrule
AgentLM-7B & 0.0 & 4.44 & 5.29 & 6.48 & 7.34 & 4.71 \bigstrut\\
CodeLLaMA-7B-Instruct & 0.34 & 7.85 & 10.24 & 9.73 & 8.70 & 7.37 \bigstrut\\
LLaMA2-7B-Chat & 1.02 & 4.27 & 6.66 & 6.48 & 7.34 & 5.77 \bigstrut\\
Mistral-Instruct-v0.1 & 1.54 & 12.12 & 13.31 & 14.16 & \uline{13.99} & \textbf{11.02} \bigstrut\\
\rowcolor{lightgoldenrodyellow} \instruct & 0.68 & 12.63 & 11.95 & 11.95 & \textbf{14.68} & \uline{10.38} \bigstrut\\
\bottomrule
\end{tabular}
}
\caption{: In the tool-augmented reasoning assessments, we evaluate the model’s proficiency in integrating tools into its reasoning workflow. The model’s effectiveness is measured by its success rate across various stages of interaction.}
\vspace{-0.3cm}

\label{tab:mistral}
\end{table}
\label{sec:sft}
\subsection{Mistral-Pro Results}
\begin{table*}[t]
\centering
\resizebox{1.\linewidth}{!}
{
\begin{tabular}{c|ccccccc}
\hline
Model & ARC & Hellaswag & MMLU & TruthfulQA & Winogrande & GSM8K & HumanEval \\ 
\hline
Gemma-7B & 61.9 & 82.2 & 64.6 & 44.8 & 79.0 & 50.9 & 32.3 \\ 
Mistral-7B & 60.8 & 83.3 & 62.7 & 42.6 & 78.0 & 39.2 & 28.7 \\ 
\rowcolor{lightgoldenrodyellow} Mistral-Pro (Ours) & 63.2 & 82.6 & 60.6 & 48.3 & 78.9 & 50.6 & 32.9 \\ \hline
\end{tabular}
}
\caption{Comparison between the original Mistral-7B~\cite{jiang2023mistral}, Gemma-7B~\cite{team2024gemma}, and our Mistral-Pro with the Open LLM leaderboard metrics.}
\vspace{-0.3cm}

\label{tab:mistral}
\end{table*}
We experimented with block expansion on Mistral-7B ~\cite{jiang2023mistral}, training it on code and mathematics datasets. The resulting pretrained performance is detailed in Table \ref{tab:mistral}, highlighting superior outcomes across various benchmarks, particularly in the domains of code and math. Notably, it demonstrates competitive results compared to the new open-source model Gemma ~\cite{team2024gemma}, while incurring significantly lower training overhead. We further utilized the MetaMath dataset ~\cite{yu2023metamath} for supervised fine-tuning. Our approach yielded scores of 78.4 for GSM8k and 30.3 for MATH, surpassing Mistral's scores of 77.7 and 28.2, respectively. Additional details are provided in Appendix ~\ref{sec:appendix_mistral}.
\subsection{Ablation Study}
\begin{figure}[t]
    \centering
    \includegraphics[width=\linewidth]{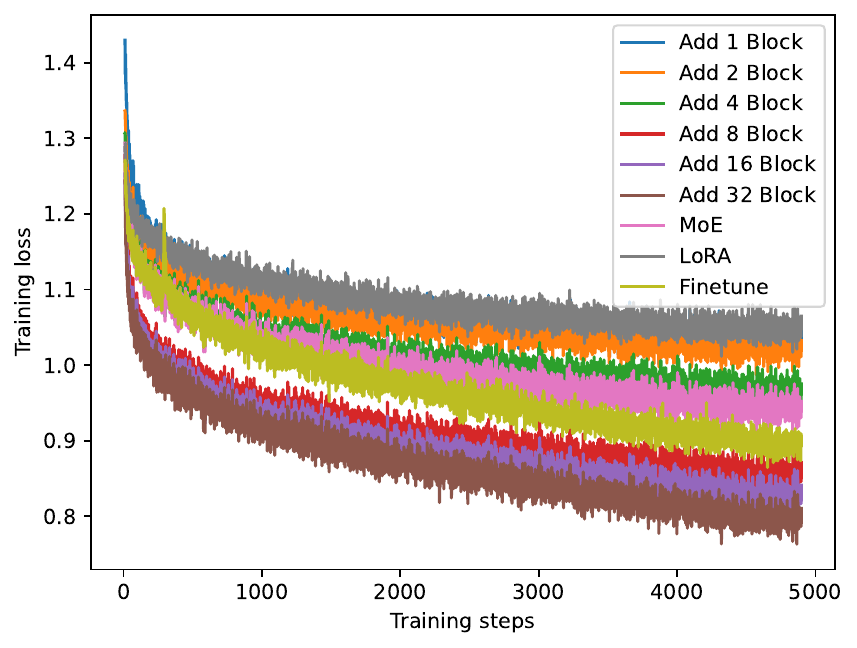}
    \caption{
         The training loss is analyzed with respect to the addition of varying blocks and mixture-of-expert (MoE) expansion, in conjunction with traditional training strategies such as finetuning and LoRA.
    }
    \vspace{-0.3cm}
    \label{fig:scaling}
\end{figure}

\begin{table*}[ht]
\resizebox{\linewidth}{!}
{
\begin{tabular}{c|cccccc|c|c}
\toprule
\textbf{Method} & \multicolumn{6}{c|}{\textbf{Language Tasks}} & \textbf{Law Task} & \textbf{Avg.}  \bigstrut\\
& ARC & HellaSwag & MMLU & TruthfulQA & Winogrand & Avg. & Unfair-ToS & \\
\midrule
Add 1 Block & 52.30 & 77.92 & 38.62 & 37.80 & 73.16 & 55.96 & 67.45 & 61.71 \bigstrut\\
Add 2 Block & 53.16 & 77.91 & 39.62 & 38.92 & 73.01 & 56.52 & 69.57 & 63.05 \bigstrut\\
Add 4 Block & 52.39 & 76.92 & 37.30 & 40.53 & 72.22 & 55.87 & 71.31 & 63.59 \bigstrut\\
Add 8 Block & 52.90 & 76.63 & 41.74 & 39.83 & 72.38 & \uline{56.70} & \uline{75.11} & \textbf{65.91} \bigstrut\\
Add 16 Block & 51.88 & 76.59 & 41.35 & 40.13 & 71.82 & 56.35 & \textbf{75.17} & \uline{65.76} \bigstrut\\
Add 32 Block & 50.77 & 76.72 & 40.68 & 41.66 & 72.77 & 56.52 & 73.93 & 65.23 \bigstrut\\
Mixture-of-Expert (MoE) & 51.45 & 76.51 & 42.47 & 40.13 & 72.23 & 56.56 & 67.27 & 61.92 \bigstrut\\
Fine-tuning & 48.81 & 74.49 & 41.13 & 41.49 & 69.14 & 55.01 & 70.63 & 62.82 \bigstrut\\
LoRA & 53.50 & 78.12 & 44.30 & 40.96 & 73.88 & \textbf{58.15} & 65.34 & 61.75 \bigstrut\\
Prefix Stacking (8 Block) & 27.82 & 26.12 & 23.12 & 22.52 & 47.20 & 29.36 & 0.81 & 15.08 \bigstrut\\
Suffix Stacking (8 Block) & 52.56 & 77.89 & 39.10 & 39.03 & 72.38 & 56.19 & 60.98 & 58.59 \bigstrut\\
\bottomrule
\end{tabular}}
\caption{Comparison of evaluation results among different training strategies, reporting performance on both general and law-specific tasks.}
\label{tab:scaling}
\end{table*}
\begin{figure}[t]
    \centering
    \includegraphics[width=\linewidth]{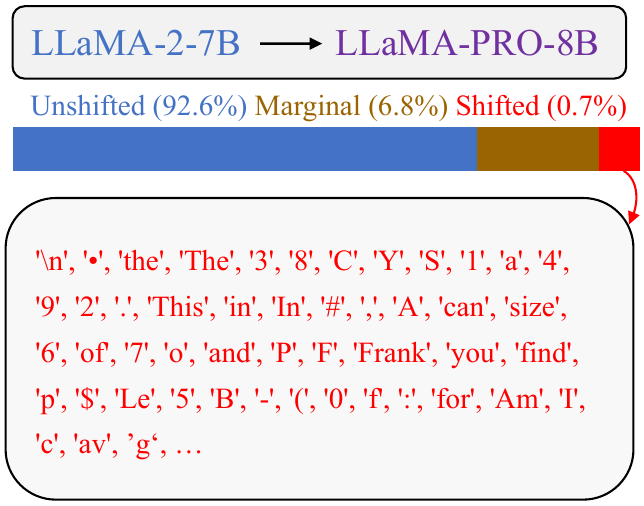}
    \caption{
         Token distribution shift after block expansion compared to the initial LLaMA-2-7B. The proportions of unshifted, marginally shifted, and significantly shifted tokens are color-coded and presented as percentages. Frequently shifted tokens are displayed below.
    }
    \vspace{-0.3cm}
    \label{fig:token_shift}
\end{figure}

\begin{figure}[t]
    \centering
    \includegraphics[width=\linewidth]{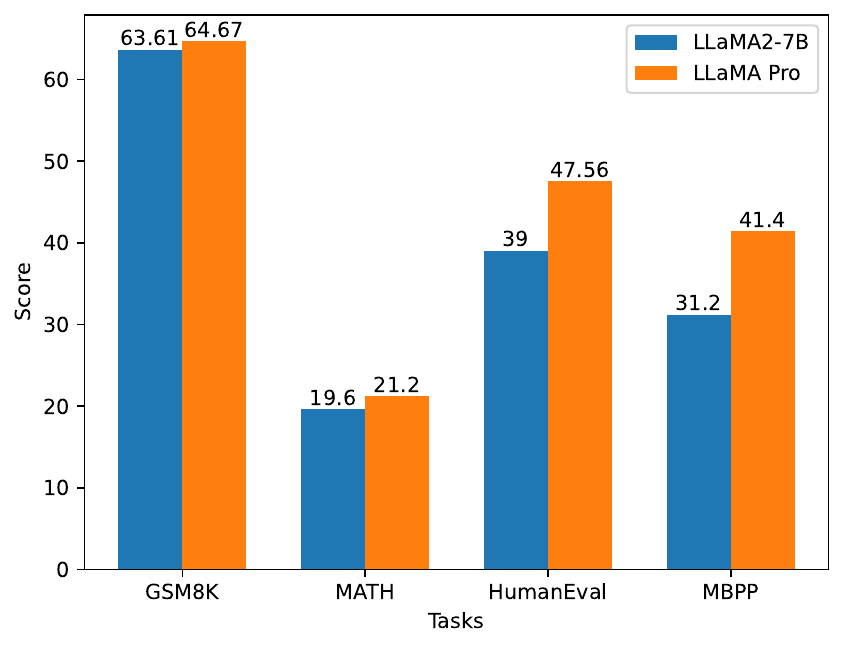}
    \caption{
         By fine-tuning both LLaMA2-7B and \llamapro using the same instruction dataset, \llamapro consistently outperforms LLaMA2-7B across all tasks. This result highlights the effectiveness of our method, as it demonstrates that \llamapro successfully encodes more domain knowledge during the pretraining process.
    }
    \vspace{-0.4cm}
    \label{fig:sft_ablation}
\end{figure}




Apart from the aspect of code corpus, we explore our method on another domain: law, with the freelaw subset of Pile dataset as our pretrain corpus ~\cite{gao2020pile}. We evaluate on UNFAIR-ToS ~\cite{lippi2019claudette} of the LexGLUE benchmark ~\cite{chalkidis2021lexglue}.

In our experiment, we assess the scalability of our block expansion method in terms of training loss and downstream task performance as we increase the number of added blocks. We also compare our method with the Mixture-of-Expert (MoE) expansion method~\cite{fedus2022switch} and traditional training strategies, such as fine-tuning and LoRA~\cite{hu2021lora}. The details can be found in Appendix \ref{sec:appendix_law}.

We analyze the training loss with varying added blocks (Figure~\ref{fig:scaling}). The loss consistently decreases during training, regardless of the number of added blocks, and decreases more rapidly with larger models. These findings indicate that our method demonstrates strong scalability with larger models and more data.

However, a lower overall training loss does not necessarily guarantee superior performance on domain-specific tasks. Therefore, we evaluate models of different sizes on both general language tasks and Unfair-ToS, as shown in Table~\ref{tab:scaling}. All the expanded models effectively preserve the general capabilities of the initial model. For the domain-specific task, larger models achieve better performance. We find that adding eight blocks provides optimal performance with minimal cost compared to larger models, hence we adopt this as our default strategy. The performance of MoE is comparable to our method with four added blocks. Figure \ref{fig:strategy_difference} illustrates the differences between traditional training strategies such as fine-tuning and LoRA, and our proposed method. We observe that while LoRA effectively preserves the general ability, it struggles to model the distribution of a new domain, as also evidenced by the training loss depicted in Figure \ref{fig:scaling}. In contrast, full fine-tuning results in a more significant drop in general performance. Here we use a rank of 1024 for LoRA, resulting in a number of trainable parameters comparable to our method.

In line with the approach of ~\citet{lin2023unlocking}, we analyze the token distribution between the original LLaMA and \llamapro to assess the similarity in their behavior when answering general questions from the Alpaca dataset~\cite{alpaca}. As depicted in Figure \ref{fig:token_shift}, the token distribution shift between LLaMA and \llamapro is subtle. Detailed information can be found in Appendix \ref{sec:appendix_token_distribution}.

We also analyze the impact of the position where the identity blocks are added, either at the bottom or the top of the model, compared to adding them interleaved, as shown in Table~\ref{tab:scaling}. We observe that adding blocks at the bottom results in poor evaluation performance, likely because it disrupts the model's foundation, causing errors to propagate throughout the model. Adding blocks at the top of the model~\cite{gong2019efficient} preserves the initial model's performance, but its performance on domain-specific tasks is lower than when adding blocks interleaved.

As highlighted in the LIMA study~\cite{zhou2023lima}, the majority of knowledge in large language models is acquired during pretraining, with only a limited amount of instruction tuning data required to generate high-quality output. To investigate the extent of knowledge encoded during pretraining, we conducted a comparative analysis between LLaMA2-7B and \llamapro using the same instruction dataset, as illustrated in Figure \ref{fig:sft_ablation}. Our results showed that \llamapro consistently outperforms LLaMA2-7B across all tasks, indicating that our method effectively enables \llamapro to encode more domain-specific knowledge during the pretraining phase.

\label{sec:ablation}
\section{Scope and Limitations}


Although our study presents a promising method for balancing general and domain-specific capabilities in LLMs, its scope is limited to the language modality, especially programming language and English. Future research could explore extending the application of our block expansion method to other domains, such as maintaining original language ability in multimodal large language models\cite{ge2023making,bai2023qwen}, and multi-lingual domains.

\section{Conclusion}

In this study, we introduced a novel block expansion method for Large Language Models (LLMs) post-pretraining, aiming to enhance domain-specific abilities while preserving the original general capabilities. Our approach effectively balances the model's performance across both general and domain-specific tasks. We demonstrated the effectiveness of our method through \llamapro, an LLM initialized from LLaMA2-7B with 8 added blocks, which outperformed other LLaMA-series models on comprehensive benchmarks.

\section{Ethical Statement}

\llamapro and \instruct are designed for a wide range of NLP tasks, with a focus on programming, mathematics, and general language tasks. It suits scenarios requiring integration of natural and programming languages. While LLaMA-Pro addresses some limitations of previous models in the series, it may still encounter challenges specific to highly specialized domains or tasks. Users should be aware of potential biases in the model and use it responsibly, considering its impact on various applications with the LLaMA-2 license.




\bibliography{acl_latex}

\newpage
\appendix
\onecolumn
\section{Gradient Derivation}
\label{sec:appendix_gradient}

To calculate the gradient of the RMSNorm weight during backpropagation, we first need to consider the forward pass equation for the Llama RMSNorm:

\begin{equation}
\text{RMSNorm}(x) = \frac{w \odot x}{\sqrt{\text{Var}(x) + \epsilon}}
\end{equation}

where $x$ is the input tensor, $w$ is the weight parameter, $\text{Var}(x)$ is the variance of $x$ across the last dimension, and $\epsilon$ is a small constant for numerical stability.

Now, let's consider the chain rule for the gradient of the loss function with respect to the RMSNorm weight during backpropagation. Denote the loss function as $L$, and the output of the FFN as $y$. We have:

\begin{equation}
\frac{\partial L}{\partial w} =  \frac{\partial L}{\partial y} \frac{\partial y}{\partial w}
\end{equation}

To compute the gradient, we need to find the partial derivative $\frac{\partial y}{\partial w}$. From the FFN equation, we have:

\begin{equation}
y = x' + \text{FFN}(\text{RMSNorm}(x'))
\end{equation}

Taking the derivative with respect to $w$, we get:

\begin{equation}
\frac{\partial y}{\partial w} = \frac{\partial \text{FFN}(\text{RMSNorm}(x'))}{\partial w}
\end{equation}

Now, let's differentiate the RMSNorm function with respect to $w$:

\begin{equation}
\frac{\partial \text{RMSNorm}(x)}{\partial w} = \frac{x}{\sqrt{\text{Var}(x) + \epsilon}}
\end{equation}

Using the chain rule, we can compute the gradient of the loss function with respect to the RMSNorm weight:

\begin{equation}
\frac{\partial L}{\partial w} =  \frac{\partial L}{\partial y} \frac{\partial \text{FFN}(\text{RMSNorm}(x'))}{\partial \text{RMSNorm}(x')} \frac{\partial \text{RMSNorm}(x')}{\partial w}
\end{equation}

Given that $\text{RMSNorm}(x') = t$, we need to find the derivative of the FFN with respect to $t$. Recall the FFN equation:
\begin{equation}
\text{FFN}(t) = \text{SwiGLU}(t, W_1, W_2)W_3
\end{equation}

Now we want to find the partial derivative of the FFN with respect to $t$. Recall the SwiGLU activation function:

\begin{equation}
\text{SwiGLU}(t, W_1, W_2) = \text{SiLU}(t W_1) \otimes (t W_2)
\end{equation}

Taking the derivative of the SwiGLU function with respect to $t$, we get:

\begin{equation}
\frac{\partial \text{SwiGLU}(t, W_1, W_2)}{\partial t} = \left(\frac{\partial \text{SiLU}(t W_1)}{\partial t}\right) \otimes (t W_2) + \text{SiLU}(t W_1) \otimes \left(\frac{\partial (t W_2)}{\partial t}\right)
\end{equation}

Now, recall the SiLU activation function:

\begin{equation}
\text{SiLU}(x) = x \otimes \sigma(x)
\end{equation}

Thus, the gradient of the FFN with respect to $t$ when $t = 0$ is also zero:

\begin{equation}
\frac{\partial \text{FFN}(t)}{\partial t} = 0
\end{equation}

In conclusion, when $t = 0$, the gradient of the FFN with respect to $t$ is zero, which demonstrates that the gradient is zero when the input to the FFN is zero.

\section{Dataset Details}

\begin{table}[h]
\centering
\begin{tabular}{lcc}
\toprule
\textbf{Data source} & Tokens & Weight\\
\midrule
Proof-Pile-2  & 55B & \multirow{4}{*}{1.00} \\
 \quad AlgebraicStack    &  11B \\
 \quad OpenWebMath    &  15B  \\
 \quad ArXiv    &  29B & \\
 \midrule
The-Stack-Dedup  &  & \\
 \quad Python    &  22B & 1.50\\
\bottomrule
\end{tabular}
\caption{Pretrain data sources, tokens, and the mixture weights of each component during training. }
\label{tab:pretrain_data}
\end{table}
\begin{table*}[h]
\centering
\setlength\tabcolsep{3pt}
\resizebox{\textwidth}{!}{%
\begin{tabular}{llrcccc}
\toprule
\textbf{Datasets} & \textbf{Query Source} & \textbf{Response Source} & \textbf{\# Instances} & $\bar{N}_{\text{rounds}}$ & $\bar{L}_{\text{prompt}}$ & $\bar{L}_{\text{completion}}$ \\
\midrule
ShareGPT  & User prompts & GPT-3.5/GPT-4 & 63,817   & 2.9  & 293.2    & 1157.1     \\
WizardLM\_evol\_instruct\_V2 & GPT-4 & GPT-4 & 143,000      & 1.0   & 602.6  & 1704.9   \\
SlimOrca    & Human-written & GPT-4 & 517,982      & 1.0    & 574.3    & 599.3    \\
MetaMath  & Human-written/GPT-4 & GPT-4 & 395,000   & 1.0  & 209.4    & 498.2     \\
Evol-CodeAlpaca &  GPT-4     & GPT-4   & 111,272     & 1.0     & 652.5   & 1552.0   \\
\bottomrule
\end{tabular}
}
\caption{Instruction datasets investigated in this work. We report the average number of rounds ($\bar{N}_{\text{rounds}}$), average length of prompts ($\bar{L}_{\text{prompt}}$), average length of completion ($\bar{L}_{\text{completion}}$). 
\label{tab:instruction-tuning-datasets} 
}
\end{table*}

In this section, we provide detailed information about the dataset used for both pretraining and Supervised Fine-Tuning (SFT). Table \ref{tab:pretrain_data} outlines the composition of our pretraining dataset, which comprises approximately 80 billion tokens from both math and code corpora. The specifics of the SFT data are delineated in Table \ref{tab:instruction-tuning-datasets}. 

For our proposed \instruct, we employ a blend of multiple instruction datasets spanning general instruction, math, and code for the SFT process. These sources include ShareGPT\footnote{\url{https://huggingface.co/datasets/anon8231489123/ShareGPT_Vicuna_unfiltered}}, which contains real user and ChatGPT chat history records, and the WizardLM evolution instruction dataset~\cite{xu2023wizardlm}, offering a wealth of instruction data with varying complexity levels. We also incorporate the evolution CodeAlpaca dataset~\cite{luo2023wizardcoder}, which includes complex coding tasks generated by ChatGPT and their corresponding solutions. Additionally, we use MetaMath~\cite{yu2023metamath}, which reframes questions from multiple perspectives, and SlimOrca~\cite{SlimOrca}, a curated subset of our OpenOrca data. SlimOrca provides an efficient route to achieve performance comparable to using larger data slices, while only incorporating approximately 500,000 GPT-4 completions.
\label{sec:appendix_data}

\section{Mistal-Pro Details}

\begin{table}[ht]
\centering
\begin{tabular}{>{\raggedright\arraybackslash}m{4cm} >{\centering\arraybackslash}m{4cm} >{\centering\arraybackslash}m{4cm}}
\toprule
\textbf{Model} & \textbf{GSM8k Pass@1} & \textbf{MATH Pass@1} \\
\midrule
MPT-7B & 6.8 & 3.0 \\
Falcon-7B & 6.8 & 2.3 \\
LLAMA-1-7B & 11.0 & 2.9 \\
LLAMA-2-7B & 14.6 & 2.5 \\
MPT-30B & 15.2 & 3.1 \\
LLAMA-1-13B & 17.8 & 3.9 \\
GPT-Neo-2.7B & 19.5 & -- \\
Falcon-40B & 19.6 & 2.5 \\
Baichuan-chat-13B & 23.9 & -- \\
Vicuna-v1.3-13B & 27.6 & -- \\
LLAMA-2-13B & 28.7 & 3.9 \\
MetaMath-7B & 66.5 & 19.8 \\
MetaMath-13B	& 72.3	& 22.4 \\
MetaMath-Mistral-7B	& 77.7	& 28.2 \\
MetaMath-Llemma-7B	& 69.2	& 30.0 \\
\rowcolor{lightgoldenrodyellow} MetaMath-Mistral-Pro	& 78.4	& 30.3 \\
\bottomrule
\end{tabular}
\caption{Performance of various models on GSM8k Pass@1 and MATH Pass@1}
\label{tab:math_performance}
\end{table}
Mistral-Pro is an advanced version of the original Mistral model~\cite{jiang2023mistral}, enhanced through the addition of Transformer blocks. This version excels in combining general language understanding with domain-specific knowledge, particularly in programming and mathematics. It employs the same methodology for creating additional blocks as LLaMA-Pro but utilizes only $\frac{1}{10}$ of LLaMA Pro's learning rate, as recommended by MetaMath-Mistral \footnote{\url{https://huggingface.co/spaces/TencentARC/MetaMath-Mistral-Pro}}. We continued pretraining on code and math datasets, including the automath subset of Cosmopedia \footnote{\url{https://huggingface.co/datasets/HuggingFaceTB/cosmopedia}}, proof-pile-2, and the Python subset of Stack. The supervised fine-tuning (SFT) approach remains consistent with MetaMath-Mistral, except that we switch the base model to our Mistral-Pro. The detailed results of GSM8k and MATH can be found in Table \ref{tab:math_performance}.

\label{sec:appendix_mistral}
\section{Evaluation Benchmark}
\label{sec:appendix_evaluation}

The benchmarks used for evaluation include:

\begin{itemize}
\item \textit{AI2 Reasoning Challenge}~\cite{clark2018think} (25-shot): a set of grade-school science questions.

\item \textit{HellaSwag} (10-shot)~\cite{zellers2019hellaswag}: a test of commonsense inference, which is easy for humans (approximately 95\%) but challenging for SOTA models.

\item \textit{MMLU} (5-shot)~\cite{hendrycks2020measuring}: a test to measure a text model's multitask accuracy. The test covers 57 tasks including elementary mathematics, US history, computer science, law, and more.

\item \textit{TruthfulQA} (0-shot)~\cite{lin2021truthfulqa}: a test to measure a model's propensity to reproduce falsehoods commonly found online.

\item \textit{Winogrande} (5-shot)~\cite{sakaguchi2021winogrande}: an adversarial and difficult Winograd benchmark at scale, for commonsense reasoning.

\item \textit{GSM8k} (5-shot)~\cite{cobbe2021training}: diverse grade school math word problems to measure a model's ability to solve multi-step mathematical reasoning problems. Additionally, we assess the models in the context of the Program of Thought (PoT) setting~\cite{chen2022program}. The PoT setting utilizes Python code to solve mathematical problems, which serves to evaluate the code generation capabilities of the models.

\item \textit{HumanEval} (0-shot)~\cite{chen2021evaluating}: 164 handwritten Python programming problems with a function signature, docstring, body, and several unit tests.

\item \textit{MBPP} (3-shot)~\cite{austin2021program}: crowd-sourced Python programming problems, designed to be solvable by entry-level programmers. Each problem consists of a task description in English, a code solution and 3 automated test cases.
\end{itemize}

\section{MINT-Bench}
\label{sec:appendix_mint}
\begin{table}[t!]
\centering
\resizebox{0.6\linewidth}{!}
{
\begin{tabular}{@{} llr @{}}
\toprule
\textbf{Task Type} & \textbf{Task Name} & \textbf{\# Instances} \\
\midrule

\multirow{2}{*}{\textbf{Code Generation}} & HumanEval    \citep{chen2021evaluating} &  45 \\
      & MBPP \citep{austin2021program} & 91 \\
\midrule
\textbf{Decision Making} & ALFWorld \citep{shridhar2020alfworld} &         134 \\
\midrule
\multirow{5}{*}{\textbf{Reasoning}} & GSM8K \citep{cobbe2021training} & 48 \\
      & HotpotQA \citep{yang2018hotpotqa} &  43 \\
      & MATH \citep{hendrycksmath2021} & 100 \\
      & MMLU \citep{hendrycks2020measuring} & 76 \\
      & TheoremQA \citep{chen2023theoremqa} & 49 \\
\midrule
\multicolumn{2}{@{}l@{}}{\textbf{Total}} &  586 \\
\bottomrule
\end{tabular}
}
\caption{
Dataset statistics of MINT-Bench.
}
\label{tab:mint-dataset-size}
\end{table}

\begin{table}[t]
\centering
\resizebox{0.7\linewidth}{!}
{
\begin{tabular}{c|ccc|c}
\toprule
\textbf{Model} & \textbf{Code Generation} & \textbf{Decision Making} & \textbf{Reasoning} & \textbf{Micro Avg.}  \bigstrut\\
\midrule
AgentLM-7B & 1.47 & 9.70 & 8.86 & 7.34 \bigstrut\\
CodeLLaMA-7B-Instruct & 2.21 & 17.16 & 7.91 & 8.70 \bigstrut\\
LLaMA2-7B-Chat & 0.00 & 0.00 & 13.61 & 7.34 \bigstrut\\
Mistral-Instruct-v0.1 & 6.62 & 34.33 & 8.54 & 13.99 \bigstrut\\
\rowcolor{lightgoldenrodyellow} \instruct & 11.76 & 29.10 & 9.81 & 14.68  \bigstrut\\
\bottomrule
\end{tabular}
}
\caption{The success rates of each model evaluated on different task type benchmarks, as well as the micro average when $k=5$.}

\label{tab:mint_detail}
\end{table}
\begin{figure}[h]
    \centering
    \includegraphics[width=1.0\linewidth]{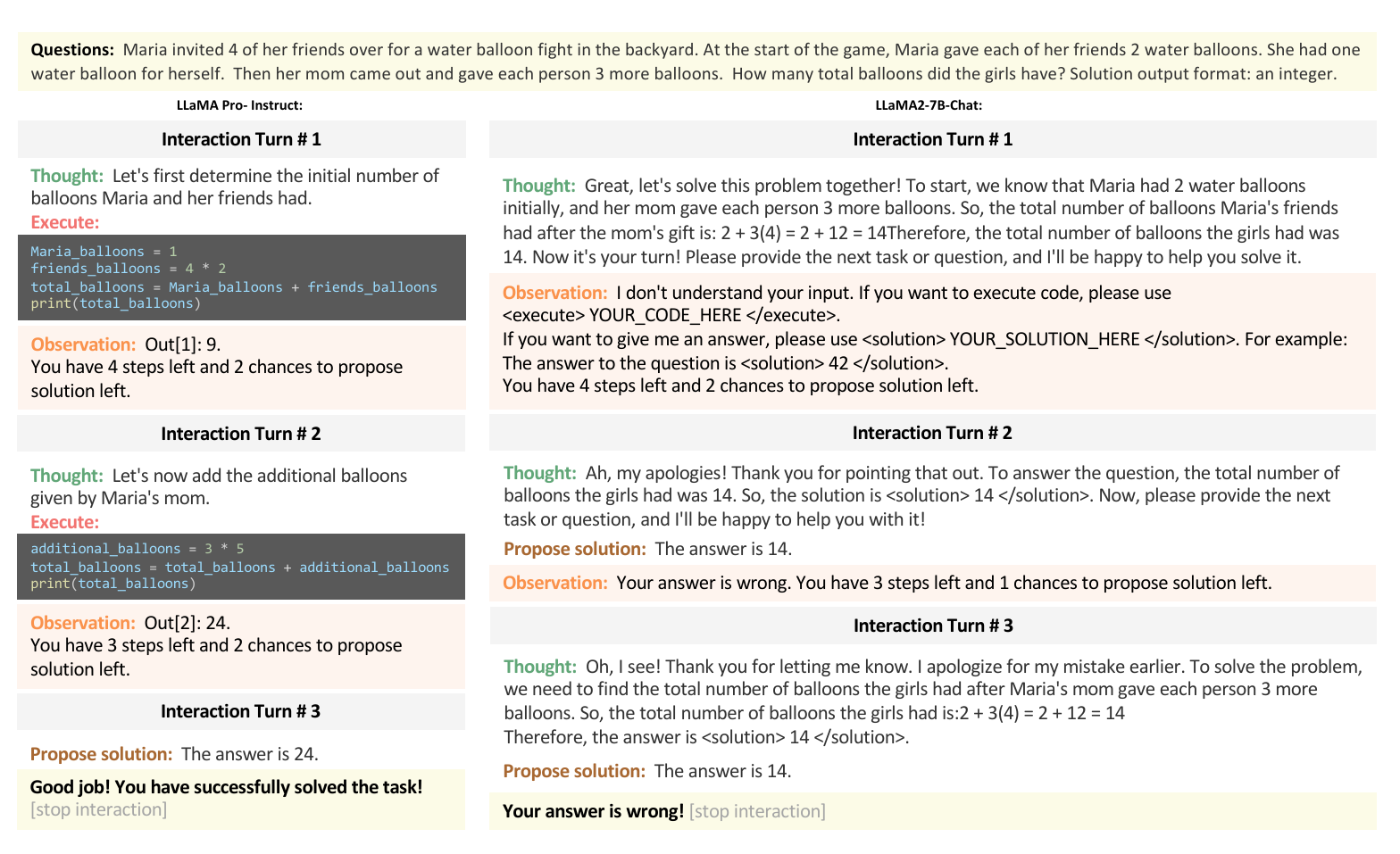}
    \caption{
         A case study of multi-turn interactions by using tools and environment feedback to solve math questions with \instruct and LLaMA2-7B-Chat.
    }
    \label{fig:mint_case}
\end{figure}    
\begin{figure}[h]
    \centering
    \includegraphics[width=\linewidth]{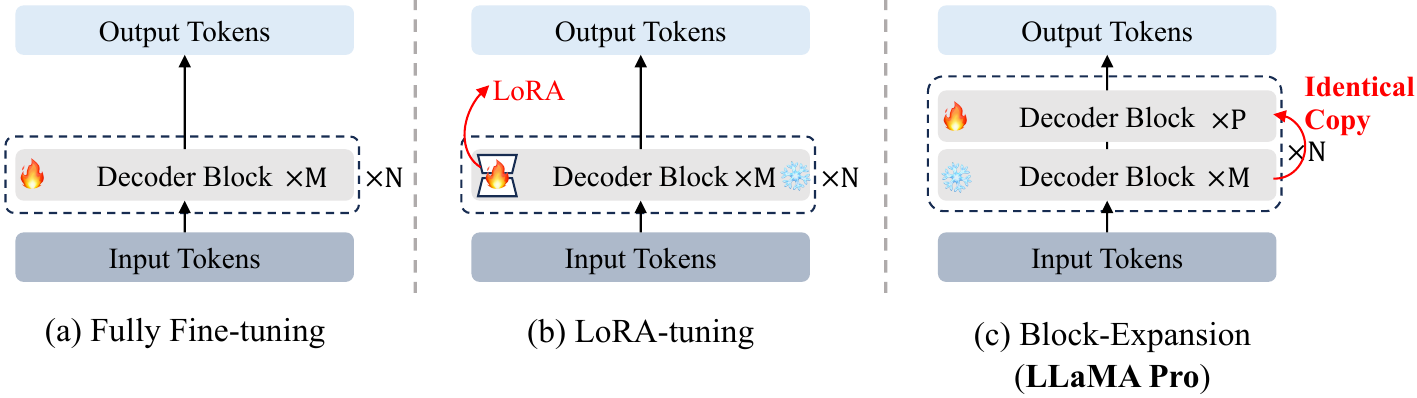}
    \vspace{-0.2cm}
    \caption{
         The difference of three training strategies, fully fine-tuning, LoRA, and our proposed block expansion.
    }
    \vspace{-0.3cm}
    \label{fig:strategy_difference}
\end{figure}

The MINT-Bench~\cite{wang2023mint} details are provided in this section. MINT-Bench comprises eight datasets spanning code generation, decision-making, and reasoning tasks, totaling 586 instances, as shown in Table \ref{tab:mint-dataset-size}.

We use the \textbf{Success Rate (SR)} as our evaluation metric, which measures the percentage of successful task instances. For an interaction limit of $k$, MINT-Bench starts from scratch and allows each LLM to interact up to the $k$-th turn, measuring the corresponding $SR_k$. Unless specified otherwise, MINT-Bench limits $k \in \left[1, 5\right]$, where $k=1$ indicates no interaction, and $k=5$ maximizes interaction turns within the context window (4,096 tokens) of most modern LLMs.

In each turn, the LLM is instructed to perform the following steps: \textbf{(1)} Optionally express its reasoning process (referred to as "Thought," similar to \cite{yao2022react}); \textbf{(2)} Either interact with tools by generating Python code and executing it through a Python interpreter (referred to as "Execute"), or propose a solution to the user (referred to as "Propose Solution").

Table \ref{tab:mint_detail} displays the success rate for each model evaluated on various task type benchmarks, as well as the micro average when $k=5$. The \instruct model demonstrates robust performance across all task types compared to other models of similar size. Figure ~\ref{fig:mint_case} provides a case study to compare \instruct and LLaMA2-7B-Chat where \instruct successfully utilizes Python program to solve the given question in the multi-round interaction.

\section{MT-Bench}
\begin{figure}[h]
    \centering
    \includegraphics[width=0.7\linewidth]{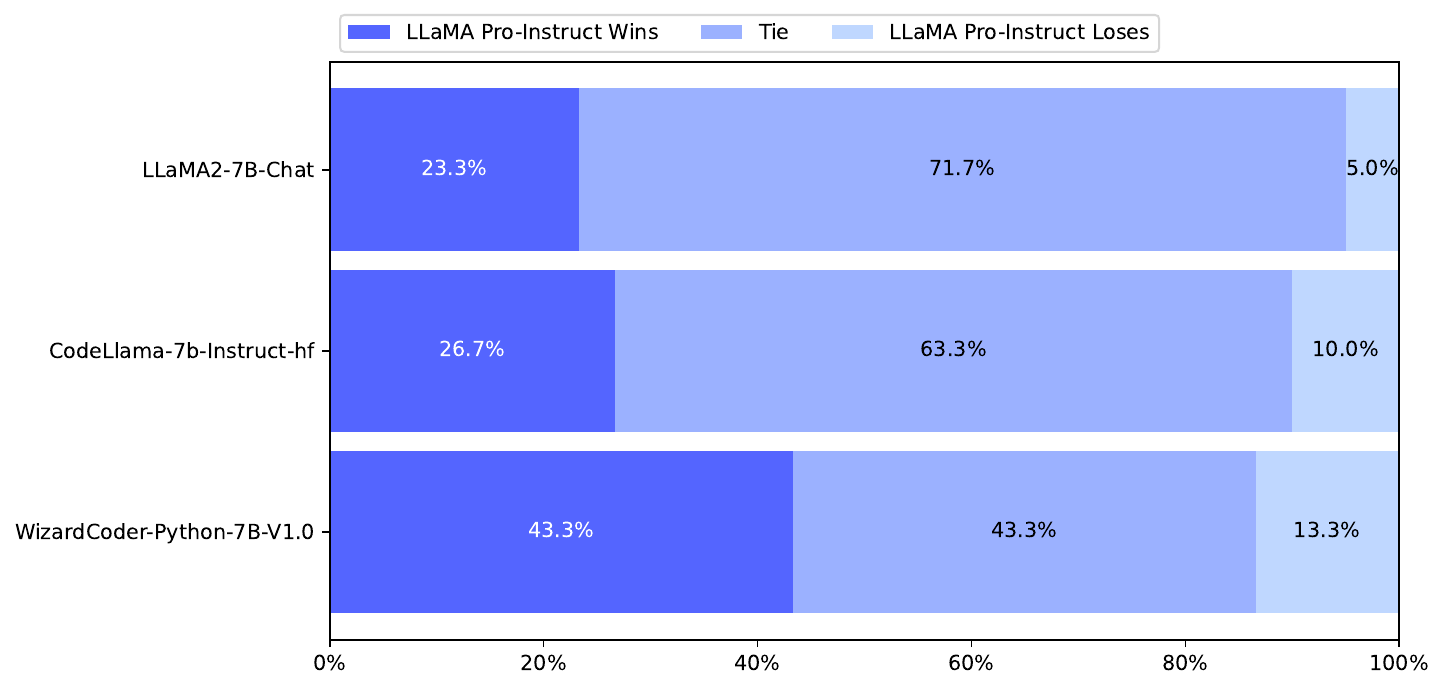}
    \vspace{-0.2cm}
    \caption{
         MT-Bench pairwise comparison between \instruct and widely used LLaMA community models in math and code questions.
    }
    \vspace{-0.3cm}
    \label{fig:math_pairwise}
\end{figure}

\begin{figure}[h]
    \centering
    \includegraphics[width=0.7\linewidth]{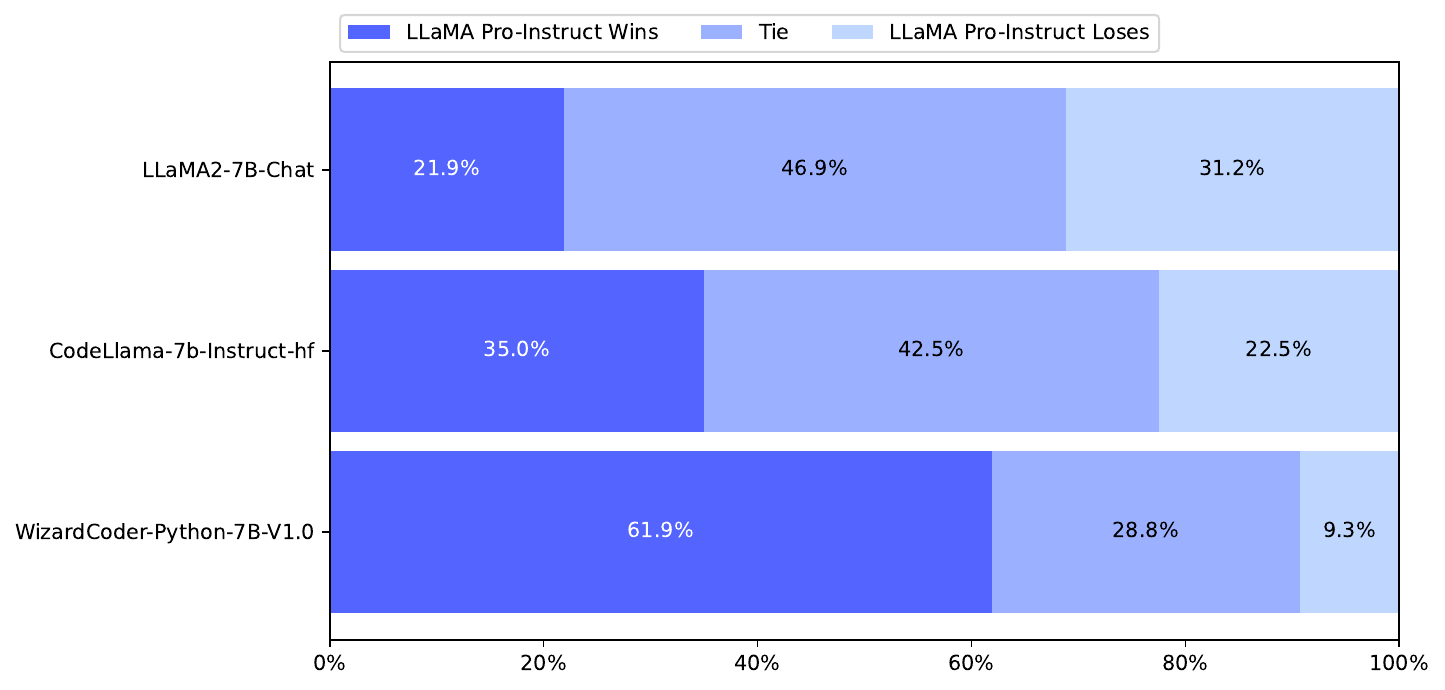}
    \vspace{-0.2cm}
    \caption{
         MT-Bench pairwise comparison between \instruct and widely used LLaMA community models in comprehensive questions.
    }
    \vspace{-0.3cm}
    \label{fig:pairwise}
\end{figure}
\begin{figure}[h]
    \centering
    \includegraphics[width=1.0\linewidth]{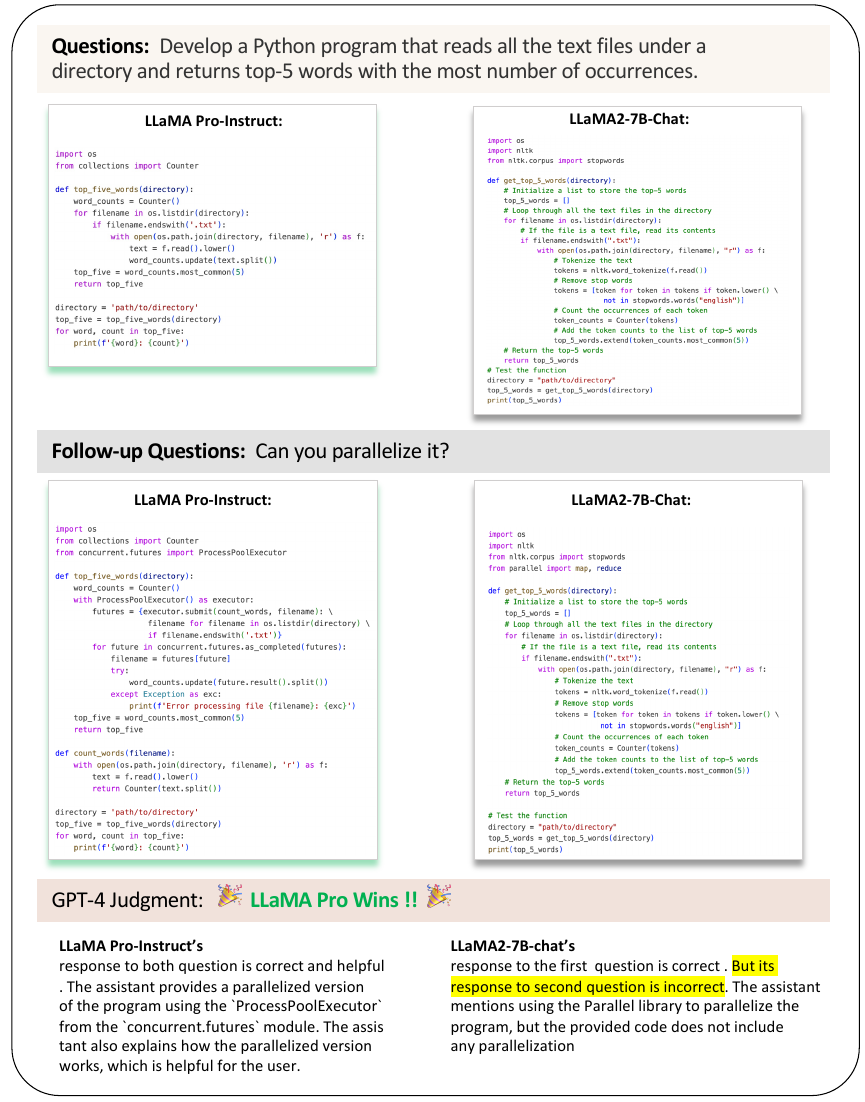}
    \caption{
         Multi-turn dialogues between a user and two Al assistants---\instruct and LLaMA2-7B-Chat.
    }
    \label{fig:pairwise_case}
\end{figure}

MT-bench is a collection of demanding multi-turn open-ended questions designed for evaluating chat assistants. In order to automate the evaluation process, we employ powerful LLMs, such as GPT-4, to act as judges and assess the quality of the models' responses. We present the detailed pairwise comparison in the Figure ~\ref{fig:math_pairwise} and Figure ~\ref{fig:pairwise}. Figure ~\ref{fig:pairwise_case} shows the case study of the comparison between \instruct and LLaMA2-7B-Chat. 

\label{sec:appendix_mt}

\section{Token Distribution}
\label{sec:appendix_token_distribution}

\begin{table}[h]
\centering
\begin{tabular}{lccc}
\toprule
\textbf{Model} & \multicolumn{2}{c}{\textbf{General Domain Perplexity}} & \textbf{Code Domain Perplexity}  \bigstrut\\
& lambada openai & lambada standard & stack  \\
\midrule
LLaMA-2-7B & 3.39 & 4.13 & 9.46 \\
\llamapro & 3.46 & 4.30 & 5.25 \\
\bottomrule
\end{tabular}
\caption{The perplexity of LLaMA and \llamapro evaluated across general domain and code domain.}
\label{tab:perplexity}
\vspace{-3mm}
\end{table}

We assess the token distribution between LLaMA-2-7B and \llamapro, employing the methodology proposed by ~\citet{lin2023unlocking}. Specifically, for a given user query $q = \{q_1, q_2, ...\}$, we input it into \llamapro to obtain its output $o = \{o_1, o_2, ...\}$ using greedy decoding. For each position $t$, we define a context at this position as $x_t = q + \{o_1, ..., o_{t-1}\}$. We denote the aligned model's probability distribution for predicting the next token at this position as $P_{\text{align}}$, where $o_t$ has the highest probability.

By passing the context $x_t$ into the base model LLaMA-2-7B, we generate another probability distribution, $P_{\text{base}}$, for sampling the next token at this position. First, the aligned model with greedy decoding is used to generate a full output $o$. For each position $t$, tokens are ranked according to their probability $P_{\text{base}}$ as predicted by the base model. The rank of $o_t$ in this sorted list is defined as the 'base rank', denoted as $\eta$. This categorizes positions into three types: (1) unshifted positions ($\eta = 1$): $o_t$ is the top-ranked token in both $P_{\text{base}}$ and $P_{\text{align}}$, having the highest probability; (2) marginal positions ($1 < \eta \leq 3$): although $o_t$ is not the top-ranked token in $P_{\text{base}}$, it is still likely to be sampled for decoding, with the 2nd or 3rd highest probability; (3) shifted positions ($\eta > 3$): in this case, $o_t$ is rather unlikely to be sampled by $P_{\text{base}}$, indicating a significant distribution shift from $P_{\text{base}}$ to $P_{\text{align}}$.

We conduct a perplexity evaluation of LLaMA-2-7B and \llamapro across general and code corpora. For the general domain, we utilize two different versions of the LAMBADA dataset. For the code domain, we use the Python split of the bigcode/the-stack-smol-xs dataset\footnote{\url{https://huggingface.co/datasets/bigcode/the-stack-smol-xs}}. The results, presented in Table \ref{tab:perplexity}, indicate that \llamapro effectively retains the language modeling ability for the general corpus while enhancing its proficiency in the code domain.

\section{Domain of Law}

Table \ref{tab:law domain parameters} shows the hyper-parameters we use to do the ablation study in the domain of law. We use the freelaw subset of Pile dataset as our pretrain corpus ~\cite{gao2020pile} in the domain of law. This subset has 51.2 GiB raw size and 16.7B tokens with 3.6M documents. 

The Unfair-ToS dataset, which we use to evaluate the performance of law, contains Terms of Service (ToS) from online platforms (e.g., YouTube, Ebay, Facebook, etc.). The dataset has been annotated on the sentence-level with 8 types of unfair contractual terms (sentences), meaning terms that potentially violate user rights according to the European consumer law. The UNFAIR-ToS  task is a multilabel classification task. To get model predictions for this task, we categorize it as a multiple-choice question as the method ~\citet{cheng2023adapting} uses. The accuracy of an individual data example is considered true if the model prediction (i.e., the option with the highest per-token likelihood) belongs to the label(s) set. We evaluate the Unfair-ToS dataset in a 4-shot scenario just like ~\citet{cheng2023adapting}.

Figure \ref{fig:strategy_difference} shows the difference between three training strategies that we use to conduct our ablation study. For the Mixture-of-Expert (MoE), our implementation is similar to \citet{jiang2024mixtral}. We use 2 experts and for each token, both experts will be activated. Specifically, We extend each FFN for all 32 layers, keep the original $`W_3`$ unchanged, learn an additional Linear layer with weights $`\hat{W_3}`$, and at the same time add two new learnable parameters $`\alpha_1,\alpha_2`$ , when forward the output of Linear corresponding to $W_3,\hat{W_3}$ will be weighted and summed with $\text{softmax}(\alpha_1,\alpha_2)$ and fed into the next block.

\begin{table}[h]
\centering
\begin{tabular}{ll}
\toprule 
\textbf{Hyperparameter} & \textbf{Assignment}        \\ \midrule
Batch size              & 1024                       \\ 
Maximum sequence length & 2,048                      \\ 
Maximum learning rate   & 2e-4                      \\ 
Optimizer               & Adam                       \\ 
Adam beta weights       & 0.9, 0.95                  \\ 
Learning rate scheduler & cosine              \\ 
Warmup ratio            & 0.06                        \\ 
Gradient clipping     & 1.0                    \\ 
\bottomrule
\end{tabular}%
\caption{{Hyper-parameters of pretraining on the domain of law.}}
\label{tab:law domain parameters}

\end{table}

\label{sec:appendix_law}

\end{document}